\pdfoutput=1
\documentclass[10pt,twocolumn,letterpaper]{article}

\usepackage{cvpr}              % To produce the CAMERA-READY version
\usepackage{graphicx}
\usepackage{amsmath}
\usepackage{amssymb}
\usepackage{booktabs}
\usepackage{multirow}
\usepackage{makecell}
\usepackage[accsupp]{axessibility}
\usepackage[pagebackref,breaklinks,colorlinks]{hyperref}

\usepackage[capitalize]{cleveref}
\crefname{section}{Sec.}{Secs.}
\Crefname{section}{Section}{Sections}
\Crefname{table}{Table}{Tables}
\crefname{table}{Tab.}{Tabs.}

\def\cvprPaperID{2587} % *** Enter the CVPR Paper ID here
\def\confName{CVPR}
\def\confYear{2023}

\begin{document}

\title{Real-time Multi-person Eyeblink Detection in the Wild for Untrimmed Video}

\author{Wenzheng~Zeng$^1$, Yang~Xiao$^1$$^\dag$, Sicheng~Wei$^1$, Jinfang~Gan$^1$, Xintao~Zhang$^1$, Zhiguo~Cao$^1$, \\Zhiwen~Fang$^{2,3}$ and Joey Tianyi Zhou$^{4,5}$ \\
 \normalsize$^1$Key Laboratory of Image Processing and Intelligent Control, Ministry of Education, School of Artificial \\ \normalsize Intelligence and  Automation, Huazhong University of Science and Technology, Wuhan 430074, China\\\normalsize$^2$School of Biomedical Engineering, Southern Medical University, Guangzhou 510515, China\\ \normalsize$^3$Department of Rehabilitation Medicine, Zhujiang Hospital, Southern Medical University, Guangzhou 510280, China\\ \normalsize$^4$Centre for Frontier AI Research, Agency for Science, Technology and Research (A*STAR), Singapore\\ \normalsize$^5$Institute of High Performance Computing, Agency for Science, Technology and Research (A*STAR), Singapore\\
\tt\small{ \{wenzhengzeng, Yang$\_$Xiao, sichengwei, jinfangan, u202115202, zgcao\}@hust.edu.cn,}\\ \tt\small fzw310@smu.edu.cn, zhouty@cfar.a-star.edu.sg
\\ \tt\small \url{https://github.com/wenzhengzeng/MPEblink}
}

\maketitle
\let\thefootnote\relax\footnotetext{\dag Yang Xiao is the corresponding author (Yang$\_$Xiao@hust.edu.cn).}
%%%%%%%%% ABSTRACT
\begin{abstract}
\vspace{-1.5mm}
Real-time eyeblink detection in the wild can widely serve for fatigue detection, face anti-spoofing, emotion analysis, etc. The existing research efforts generally focus on single-person cases towards trimmed video. However, multi-person scenario within untrimmed videos is also important for practical applications, which has not been well concerned yet. To address this, we shed light on this research field for the first time with essential contributions on dataset, theory, and practices. In particular, a large-scale dataset termed MPEblink that involves 686 untrimmed videos with 8748 eyeblink events is proposed under multi-person conditions. The samples are captured from unconstrained films to reveal ``in the wild" characteristics. Meanwhile, a real-time multi-person eyeblink detection method is also proposed. Being different from the existing counterparts, our proposition runs in a one-stage spatio-temporal way with end-to-end learning capacity. Specifically, it simultaneously addresses the sub-tasks of face detection, face tracking, and human instance-level eyeblink detection. This paradigm holds 2 main advantages: (1) eyeblink features can be facilitated via the face's global context (e.g., head pose and illumination condition) with joint optimization and interaction, and (2) addressing these sub-tasks in parallel instead of sequential manner can save time remarkably to meet the real-time running requirement. Experiments on MPEblink verify the essential challenges of real-time multi-person eyeblink detection in the wild for untrimmed video. Our method also outperforms existing approaches by large margins and with a high inference speed.
% Extensive experiments verify the superiority of the proposed method in both effectiveness and efficiency, while also highlighting the essential challenges of the proposed task.

% The dataset and source code can be found in 
% \url{https://github.com/wenzhengzeng/MPEblink}.
\end{abstract}

%%%%%%%%% BODY TEXT
\vspace{-6mm}

\begin{figure}[t]
\begin{center}
%\fbox{\rule{0pt}{2in} \rule{0.9\linewidth}{attention.pdf}}
\includegraphics[width=0.48\textwidth]{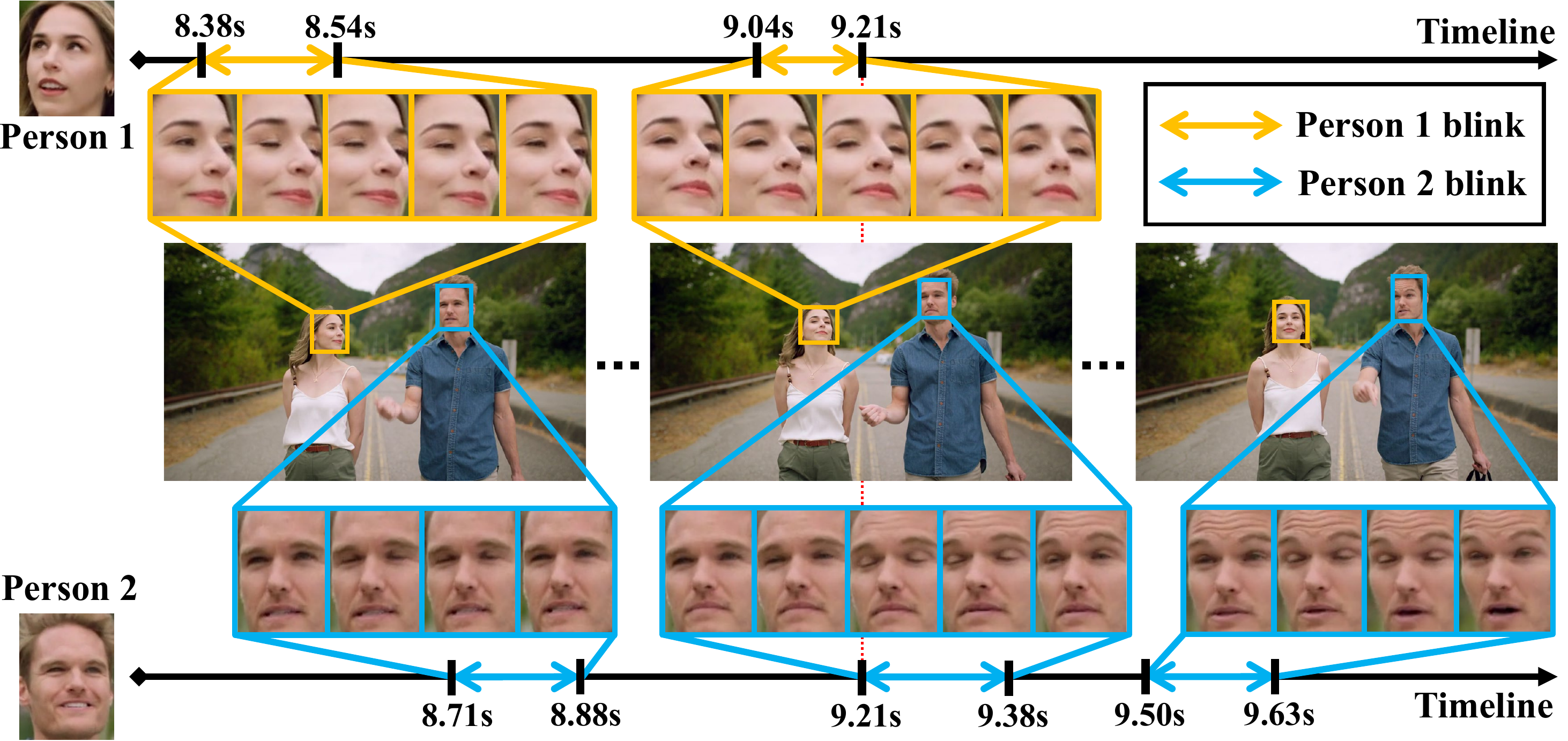}
\vspace{-9.5mm}
\end{center}
        \caption{The illustration on multi-person eyeblink detection. This task aims at being aware of the presence of people and detecting their eyeblink activities at instance level.}
\label{fig:task_formulation}
\vspace{-3mm}
\end{figure}

\section{Introduction}
\vspace{0.1mm}

Real-time eyeblink detection in the wild is a recently emerged challenging research task~\cite{eyeblink_Hu} that can widely serve for fatigue detection~\cite{2_drive_fatigue_detection}, face anti-spoofing~\cite{dataset_zju_ICCV_2007}, affective analysis~\cite{eyeblink_applications}, etc. Although remarkable progress has been made~\cite{eyeblink_Hu, eyeblink_mEBAL_2020, eyeblink_alebk}, the existing methods generally focus on single-person cases within trimmed videos. Multi-person scenario within untrimmed videos has not been well concerned yet. However, detecting long-term eyeblink behaviors at multi-instance level is more preferred for some practical application scenarios. For example, it can be used to estimate attendees' attention level and emotional state change during social interaction~\cite{eyeblink_social, eyeblink_alebk,eyeblink_applications}. Thus, effective and real-time multi-person eyeblink detection in the wild for untrimmed video is indeed required.

% are proposed without concerning eyeblinks in multi-person scenarios towards untrimmed videos. However, towards some practical application scenarios detecting eyeblink

% Some efforts~\cite{eyeblink_Hu, eyeblink_mEBAL_2020} have already been paid to this. Nevertheless, most of them are proposed without considering eyeblinks in multi-person scenarios towards untrimmed videos. Meanwhile, the existing eyeblink detection in the wild dataset~\cite{eyeblink_Hu} generally focus on single person towards trimmed short video clips. However, towards some practical application scenarios detecting eyeblink in multi-person scenarios within untrimmed long videos is more preferred. For instance, it is of great value for estimating attendants' attention level and emotional state change from their eyeblink behaviors during social interaction~\cite{eyeblink_mEBAL_2020}. 
% % In this case, the algorithm not only requires detecting eyeblink events on temporal axis, but also needs to track instances across frames to obtain a complete awareness of each instance's behavior, which is much more challenge than existing formulations. 
% % In this case, the algorithm needs to have a complete awareness of each instance along the whole video and detect eyeblinks at instance level, which is much more challenging than previous formulations.
% Thus, effective and real-time multi-person eyeblink detection in the wild for untrimmed video is essentially required to ensure the performance.

To this end, we shed the light on this research problem with essential contributions on dataset, theory, and practices. First, a challenging labeled multi-person eyeblink detection benchmark termed MPEblink is built under in-the-wild conditions. It consists of 686 untrimmed long videos captured from unconstrained movies to reveal the ``in the wild" characteristics. The contained scenarios are realistic and diverse, such as social interactions and group activities. To our knowledge, MPEblink is the first multi-person eyeblink detection dataset that focuses on in-the-wild long videos.~\cref{fig:task_formulation} illustrates a sample video with ground truth annotations within it. Different from the existing eyeblink detection benchmarks~\cite{dataset_eyeblink8, eyeblink_mEBAL_2020, eyeblink_Hu}, the proposed benchmark aims at being aware of all the attendees along the whole video and detecting their eyeblinks at instance level. In summary, MPEblink is featured with multi-instance, unconstrained, and untrimmed, which makes it more challenging and realistic than previous formulations. 
% This new benchmark opens up possibilities for applications of eyeblink detection at multi-instance level.

To perform eyeblink detection, previous methods~\cite{eyeblink_Hu,dataset_eyeblink8, eyeblink_mEBAL_2020, eyeblink_alebk, eyeblink_landmark} generally take a sequential approach containing face detection, face tracking, and classification on the pre-extracted local eye clues within a temporal window. Whereas such a pipeline seems reasonable, it has several critical drawbacks. First, the use of isolated components may lead to sub-optimal results as they are not jointly optimized and it is inefficient due to the inability to reuse the features of each stage. 
% A derived critical defect is that 
Second, the eyeblink features only contain the information from the pre-extracted local eye clues (i.e., a small part of face), lacking the useful information from global face context such as head pose and illumination condition that are crucial for detecting eyeblinks in the wild. Moreover, the pre-extracted local eye clues are also unreliable due to the localization challenge towards unconstrained scenarios. Third, the sequential approach leads to computational cost being sensitive to the subject amount, which is hard to meet the real-time running requirement.

To tackle these issues, we propose a one-stage multi-person eyeblink detection framework called InstBlink, which can simultaneously detect human faces, track them over time, and do eyeblink detection at instance level. InstBlink takes inspiration from the existing query-based methods~\cite{sparsercnn, tuber,tevit, efficientvis} and models the spatio-temporal face as well as eyeblink representations at instance level within each query. The insight is that the features can be effectively shared among these sub-tasks and the eyeblink features can be facilitated via face's global contexts (e.g., head pose and illumination condition) with joint optimization and interaction, 
% which is essentially helpful for in the wild scenarios. 
especially in unconstrained in-the-wild cases. 
Experiments verify the superiority of InstBlink in both effectiveness and efficiency, while also highlighting the critical challenges of real-time multi-person eyeblink detection in the wild for untrimmed video.

% Under such a synergy, face detection and tracking capacity can also be facilitated by eyeblink clues. Experiments verify the superiority of Instblink in both effectiveness and efficiency, while also highlighting the essential challenges of the proposed task.

% Our method is a much simplified architecture and with high running speed (i.e., 104 FPS). 

% Experimental results clearly demonstrate the advantage of our new algorithm and reveal insights for future improvement.

The main contributions of this work lie in 3 folders:

$\bullet$ To our knowledge, it is the first time that instance-level multi-person eyeblink detection in untrimmed videos is formally defined and explored;

% $\bullet$ An unconstrained multi-person eyeblink detection dataset MPEblink that contains 686 untrimmed videos with 8748 eyeblink events is built;

$\bullet$ We introduce an unconstrained multi-person eyeblink detection dataset MPEblink that contains 686 untrimmed videos with 8748 eyeblink events, featured with more realistic and challenging;

$\bullet$ We propose a one-stage multi-person eyeblink detection method that can jointly perform face detection, tracking, and instance-level eyeblink detection. Such a task-joint paradigm can benefit the sub-tasks uniformly.

% \begin{itemize}
% \vspace{-3mm}
% \item To our best knowledge, it is the first time that instance-level multi-person eyeblink detection in untrimmed video is formally defined and explored.
% \vspace{-3mm}
% \item We introduce an unconstrained multi-person eyeblink detection dataset which contains 686 untrimmed videos and 8748 eyeblink events, featured with more realistic and challenging.
% \vspace{-3mm}
% \item We propose an one-stage multi-person eyeblink detection framework that can jointly perform instance detection, tracking and their eyeblink detection. Such task-joint paradigm can benefit the sub-tasks uniformly. 
% % \item Our method outperforms other state-of-the-art approaches by large margins, and with high real-time running speed.
% \end{itemize}

\section{Related Work}
\noindent \textbf{Eyeblink detection dataset.}
Existing eyeblink detection datasets~\cite{dataset_eyeblink8, dataset_research_night, dataset_silesian5, dataset_talkingface,dataset_zju_ICCV_2007,eyeblink_mEBAL_2020} generally focus on constrained indoor cases with a consistent environmental setup. HUST-LEBW~\cite{eyeblink_Hu} extends the scenarios to the unconstrained outdoor cases. From then on, the community starts to pay more attention to eyeblink detection in the wild. Nevertheless, HUST-LEBW mainly focuses on single-person scenarios towards trimmed videos, which limits the application of methods in more realistic scenarios such as group activities and social interactions. To address such limitation, we introduce a new dataset termed MPEblink. The proposed dataset is featured with multi-person, unconstrained, and untrimmed, thus being more realistic and challenging.

\noindent \textbf{Eyeblink detection method.}
Existing methods generally judge eyeblinks from the pre-extracted local eye clues (e.g., local eye region~\cite{eyeblink_Hu, eyeblink_alebk, eyeblink_mEBAL_2020, eyeblink_RTBENE,deep_fake_eyeblink} or landmarks around eyes~\cite{eyeblink_landmark, eyeblink_unsupervised_2020, blinkdetection+, dataset_rldd}). To obtain the local eye clues, the existing works generally run in a sequential way including face detection, face tracking, and landmark detection. Finally, the eyeblink result is classified from the pre-extracted local eye clues within a temporal window. 
Such a pipeline is inefficient due to the isolated optimization among different sub-tasks. Meanwhile, the eyeblink features only contain the information from the pre-extracted local eye clues, lacking useful global information such as head pose and illumination, and the pre-extracted eye clues are also unreliable due to the localization challenge toward unconstrained cases. Besides, due to the multi-stage 
characteristic, existing methods do not scale well with the number of people in the scene to meet the real-time running requirement.
In this paper, we propose a one-stage eyeblink detection framework that can simultaneously detect human faces, track them along the time, and do eyeblink detection at instance level. Within it, eyeblink features can be facilitated via global face context, and the features can be effectively shared among the sub-task to achieve high inference speed.

\begin{figure*}[t]
\begin{center}
%\fbox{\rule{0pt}{2in} \rule{0.9\linewidth}{attention.pdf}}
\includegraphics[width=0.85\textwidth]{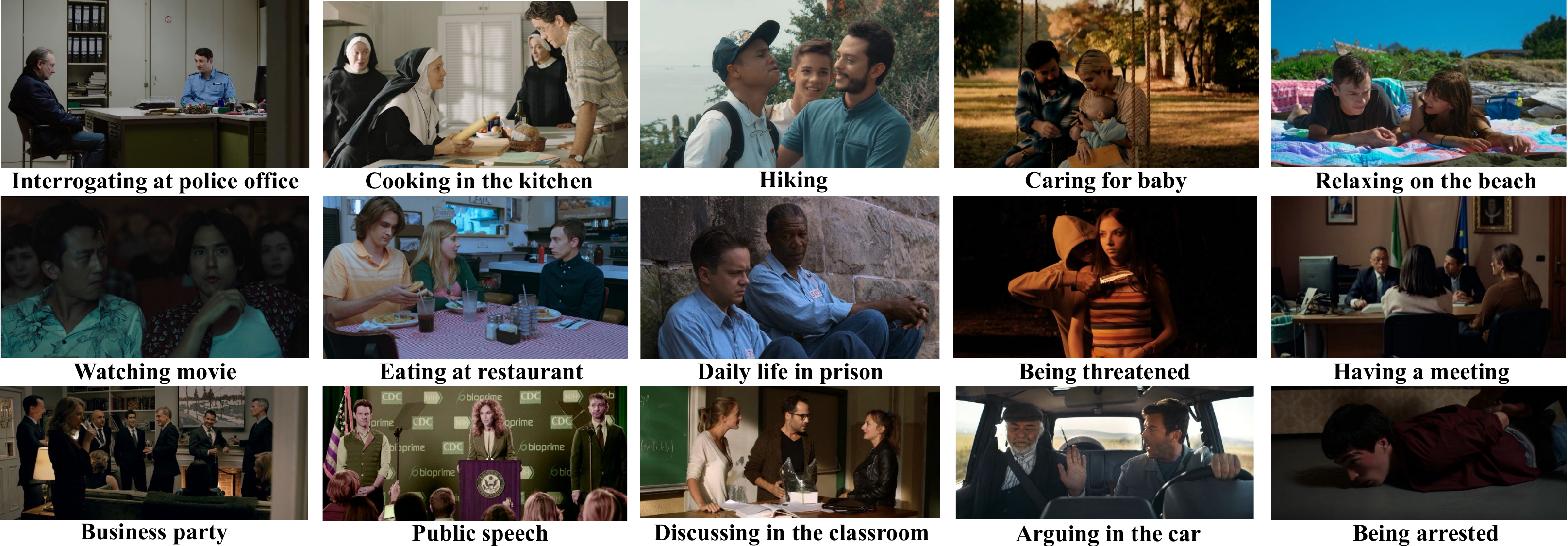}
\vspace{-5mm}
\end{center}
        \caption{The snapshots within MPEblink. Our dataset is featured with more realistic and diverse scenarios.}
\label{fig:snapshots}
\vspace{-5mm}
\end{figure*}

\noindent \textbf{Spatio-temporal action detection.}
Spatio-temporal action detection~\cite{jhmb, ucf101, ava, multisports, act_detector, moc, tuber} aims at simultaneously obtaining the spatial and temporal location of an action, which is defined as an action tube~\cite{act_detector}. The task only focuses on the isolated action tubes without instance awareness and thus can not make an instance-level analysis along the whole video. Different from that, our proposed multi-person eyeblink detection task aims at simultaneously being aware of the presence of people and analyzing their eyeblink behaviors at instance level. 
% We believe our formulation can also benefit the field of general spatio-temporal action detection.

\noindent \textbf{Query-based methods.} Query-based methods are impacting many computer vision tasks. After DETR~\cite{detr}, the pioneer that introduces the query-based framework to address object detection, numerous efforts have been subsequently paid to facilitate visual recognition under the query concept in the fields of object detection~\cite{deformabledetr, sparsercnn, conditionaldetr, dndetr}, pose estimation~\cite{petr}, action detection~\cite{tadtr,tuber}, and segmentation~\cite{mask2former,vistr,efficientvis,tevit}. 
Inspired by these, we build 
the first one-stage multi-person eyeblink detection framework for untrimmed video. We propose a unified solution to simultaneously model the instance-level face and eyeblink representations in the whole video.

\section{The MPEblink Benchmark}

Existing eyeblink detection datasets generally focus on single-person scenarios, and are also limited in aspects of constrained conditions or trimmed short videos. To explore unconstrained eyeblink detection under multi-person and untrimmed scenarios, we construct a large-scale multi-person eyeblink detection dataset termed MPEblink to shed the light on this research topic that has not been well studied before. The distinguishing characteristics of MPEblink lie in 3 aspects: multi-person, 
unconstrained, and untrimmed long video, which makes it more realistic and challenging than previous datasets.
% ~\cref{sec:problem_definition} illustrates our task formulation. The data collection, annotation, and its statistics will be introduced in ~\cref{sec:data_collectiong},~\cref{sec:data_annotation}, and~\cref{sec:data_statistics} respectively. The newly proposed evaluation metrics will be introduced in~\cref{sec:evaluation_metrics}.

\subsection{Task Formulation} \label{sec:problem_definition}
We think that a good multi-person eyeblink detection algorithm should be able to (1) detect and track human instances' faces reliably to ensure the instance-level analysis ability along the whole video, and (2) detect eyeblink boundaries accurately within each human instance to ensure the precise awareness of their eyeblink behaviors.

Formally, for an untrimmed video with $T$ frames and $N_{gt}$ people in it, towards the $j$-th person, let $\hat{l}^{j}\in \mathbb{R}^{T \times 4}$ denote its face bounding boxes across the video. $\hat{c}^j\in \mathbb{R} ^{T}$ is given as the face labels to reflect the face existence of this instance in each frame (i.e., $\hat{c}_t^j\in \left\{ 1,0 \right\}$, where $\hat{c}_t^j=0$ indicates that the face is not visible due to serious occlusion or departure at time $t$). $\hat{l}_t^j$ is also set to $\varnothing$ when the person is not 
visible. We use $\hat{l}^j$ and $\hat{c}^j$ to represent the instance-level information of each person. Suppose person $j$ blinks $\hat{K}^j$ times in the video. Let $\hat{B}_{\hat{k}}^{j}=\left[ \hat{s}_{\hat{k}}^{j}, \hat{e}_{\hat{k}}^{j} \right]$ denote the $\hat{k}$-th eyeblink event within this person, where $\hat{s}_{\hat{k}}^{j}$ and $\hat{e}_{\hat{k}}^{j}$ 
denote its starting and ending time. Suppose a multi-person eyeblink detection algorithm produces $H$ human instance hypotheses. Within each human instance hypothesis $i$, it needs to produce a sequence of predicted face bounding boxes $l^i$, face classification scores $c^i$, and $K^i$ eyeblink proposals $B_{k}^{i}=\left[{s}_{k}^{i},e_{k}^{i} \right]$. Better performance can be acquired when predictions get closer to ground truths.

\begin{table*}[t]
\setlength\tabcolsep{3pt}
\centering
  \scriptsize
  \caption{High-level statistics of MPEblink and existing eyeblink detection datasets. s: seconds, m: minutes. f: frames.}
  \vspace{-3mm}
\begin{tabular}{c|cccccccc}
\toprule
                & Talking Face~\cite{dataset_talkingface} & ZJU~\cite{dataset_zju_ICCV_2007}        & Eyeblink8~\cite{dataset_eyeblink8} & Silesian5~\cite{dataset_silesian5} & Researcher's Night~\cite{dataset_research_night} & mEBAL~\cite{eyeblink_mEBAL_2020} & HUST-LEBW~\cite{eyeblink_Hu} & MPEblink (Ours)      \\ \midrule
Num. of videos  & 1            & 80         & 8         & 5         & 107                & 6000  & 673 / 90       & 686                  \\
Video length    & 3.3m         & 4.1-4.7s & 2.7-8.9m  & 1.4-2.7m  & 0.3-3.4m           & 0.6s & 13f / 0.7-49.4s     & 7.1-85.9s           \\
Unconstrained   & $\times$        & $\times$        & $\times$     & $\times$      & $\times$              & $\times$    & \checkmark         & \checkmark                    \\
Instances/video & 1            & 1          & 1         & 1         & 1                  & 1     & 1         & 1-8 \\
Blink count     & 61           & 255        & 353       & 300       & 1867               & 3000  & 381 / 176       & 8748 \\ \bottomrule               
\end{tabular}
\label{tab: data statistics}%
\vspace{-5mm}
\end{table*}

\subsection{Data Collection} \label{sec:data_collectiong}

% We collect 686 untrimmed video clips with various length (i.e., 7.1-85.9s) from 86 different unconstrained movies that involve the "in the wild" characteristics. The number of human instance in each video varies from 1 to 8. Some snapshots within the dataset is shown in~\cref{fig:graphical_statistics}. It can be seen that the samples are of diverse scenarios (e.g, social interactions at classroom and party, or during hiking and driving, etc). Besides, there also exist many challenging cases such as occlusion, motion blur, people in and out, bad illumination and extreme pose, which makes our dataset more challenging and close to realistic scenarios. It is worthy noting that the collected videos are also of high resolution (e.g, 1080P or 4K) than 
% previous arts, which can preserve details of eyes and enable fine-grained feature extraction or analysis for future research.
\begin{figure}[t]
\begin{center}
%\fbox{\rule{0pt}{2in} \rule{0.9\linewidth}{attention.pdf}}
\includegraphics[width=0.45\textwidth]{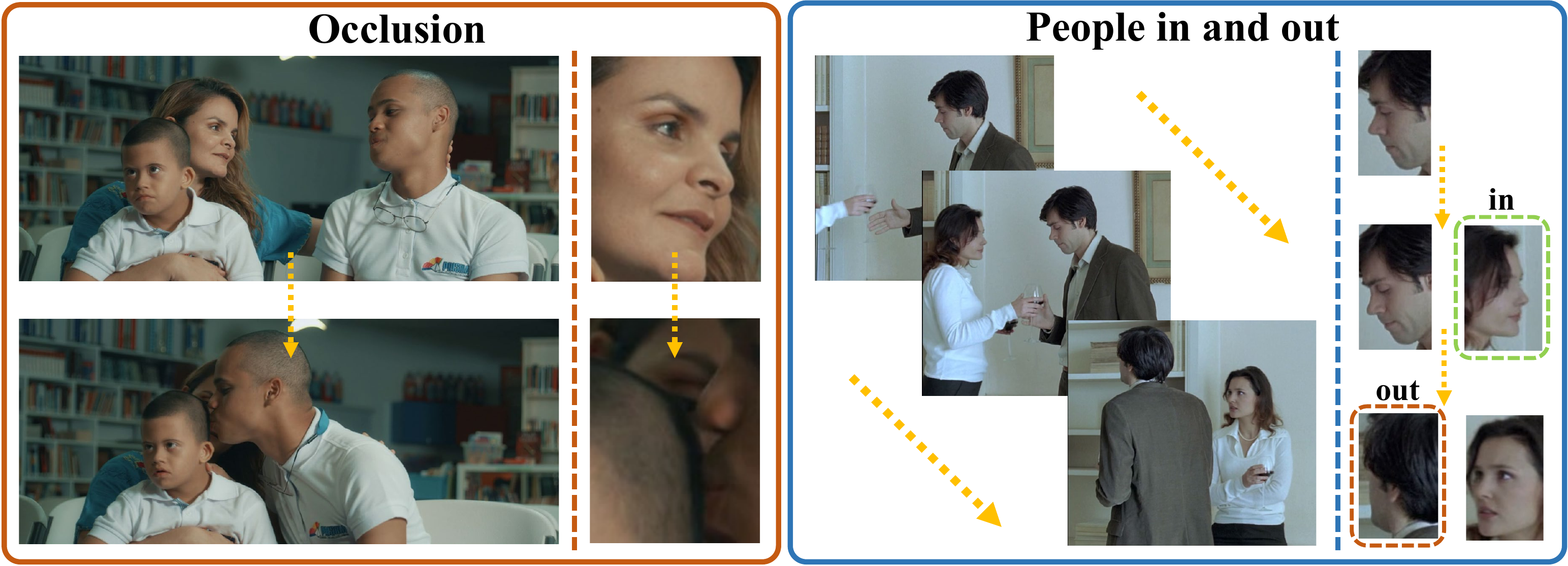}
\vspace{-5mm}
\end{center}
        \caption{Occlusion and people in \& out cases during interaction among instances.
}
\label{fig:challenges}
\vspace{-2mm}
\end{figure}

\begin{figure}[t]
\begin{center}
%\fbox{\rule{0pt}{2in} \rule{0.9\linewidth}{attention.pdf}}
\includegraphics[width=0.45\textwidth]{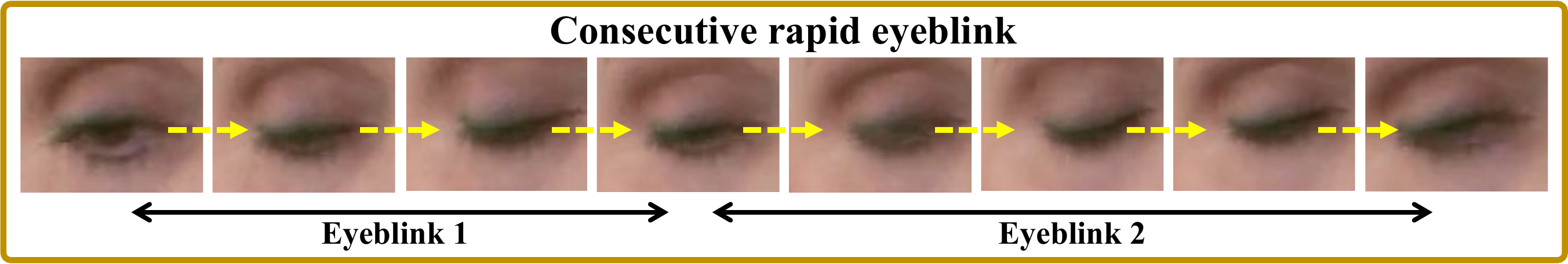}
\vspace{-5mm}
\end{center}
        \caption{An example of rapid and consecutive eyeblinks in untrimmed video.
}
\label{fig:consecutive_eyeblink}
\vspace{-5mm}
\end{figure}
We collect 686 untrimmed video clips with various lengths (i.e., 7.1-85.9s) from 86 different unconstrained movies that involve the ``in the wild” characteristics. Some snapshots of the acquired samples are shown in~\cref{fig:snapshots}. It can be observed that the samples are with variational indoor and outdoor scenarios for different themes. 
Thus, the acquired samples from these movies are much more realistic and closer to practical applications. 
Meanwhile, due to the high divergence of the selected movie data source, the captured multi-person eyeblink samples are also of great challenges for effectively detecting eyeblinks. For instance, the pose, illumination condition, expression, and face size among instances differ a lot, even within the same video. Moreover, because of the various types of interaction among instances, there may exist severe occlusion or people in and out scenarios as shown in~\cref{fig:challenges}, making it harder for both instance localization and their eyeblink event detection.
Besides, different from the existing unconstrained counterpart~\cite{eyeblink_Hu} that mainly focuses on trimmed videos, MPEblink targets on untrimmed setting.
As a result, we not only need to recognize eyeblink but also need to accurately localize the starting and ending time of each eyeblink event, which is more challenging especially when rapid and consecutive eyeblinks occur.
% As a result, the main goal has shifted from recognizing eyeblinks to accurately localizing the starting and ending time of each eyeblink event, which is more challenging especially when rapid and consecutive eyeblinks occur.
An example is shown in~\cref{fig:consecutive_eyeblink}. It can be observed that the appearance difference among eye statuses is small when consecutive rapid eyeblinks happen, making it hard to distinguish the boundary of each eyeblink.
Considering the above challenges together, it can be summarized that unconstrained multi-person eyeblink detection in untrimmed videos is indeed a challenging research task.

% It is worthy noting that the collected videos are also of higher resolution (e.g, almost 1080P) than 
% previous arts, which can preserve details of eyes and enable fine-grained feature extraction or analysis for future research.

\subsection{Data Annotation} \label{sec:data_annotation}
For human instances in each video, we labeled their face bounding boxes exhaustively across the whole video. To facilitate the research of landmark-based methods and include a broader range of applications, 68 facial landmark positions~\cite{fake_it_till_you_make_it} are also annotated. Technically, this part of the annotation is under a semi-supervised manner: we use the state-of-the-art face analysis engine InsightFace~\cite{insightface} to achieve face bounding box~\cite{scrfd} and landmark~\cite{fake_it_till_you_make_it} detection. To track instances across frames, a matching strategy that considers bounding box IoU and similarity among deep face features~\cite{arcface} is designed. Besides, human annotators carefully check the annotation quality and correct the wrong bounding boxes and tracking results generated by the algorithm. For eyeblink events within each instance, human annotators carefully define their start and end frame. Finally, 8748 eyeblink events are labeled. 
% The annotations are cross-validated by 2 annotators, which should be inspected and revised if they differ more than 1 frame in total. Thus, our annotation is of high quality.

\subsection{Data Statistics}\label{sec:data_statistics}

The dataset statistics comparison among MPEblink and the other eyeblink detection datasets are given in Table~\ref{tab: data statistics}. Actually, MPEblink provides the largest number of eyeblink events (i.e., 8748) within 686 untrimmed videos. The biggest difference from the previous datasets is that the videos in our dataset contain various number of human instances (i.e., 1-8) with exhaustive annotation. The videos are captured under unconstrained in-the-wild conditions with high diversity as shown in~\cref{fig:snapshots}. Moreover, existing in-the-wild dataset~\cite{eyeblink_Hu} mainly focuses on trimmed cases and only provides a limited number (i.e., 90) of untrimmed videos for testing purposes, while MPEblink targets on untrimmed cases, which is superior in both quantity and diversity. Overall, MPEblink is featured with more realistic and challenging and thus has broader application values.
% Moreover, the collected videos are also of high resolution (almost at 1080P or 4K) than previous arts, which can preserve details of eyes and enable fine-grained feature extraction or analysis for future research. 

\begin{figure*}[t]
\begin{center}
%\fbox{\rule{0pt}{2in} \rule{0.9\linewidth}{attention.pdf}}
\includegraphics[width=0.81\textwidth]{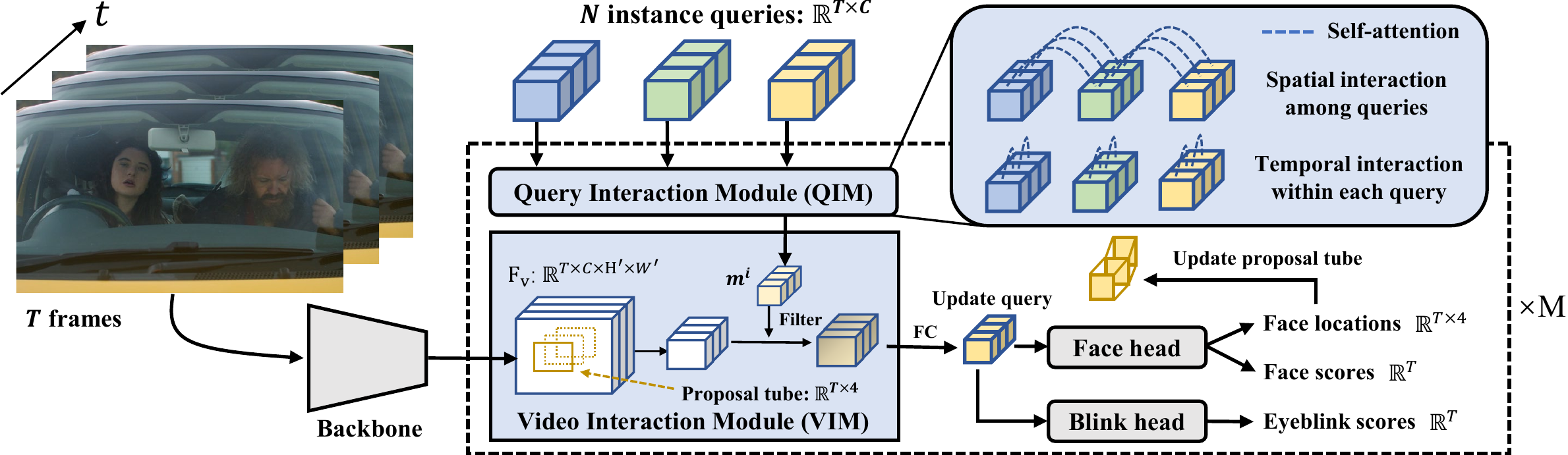}
\vspace{-5mm}
\end{center}
        \caption{Overview of the InstBlink framework.}
\label{fig:pipeline}
\vspace{-5mm}
\end{figure*}

\subsection{Evaluation Metrics}\label{sec:evaluation_metrics}
Existing metrics for eyeblink detection only focus on single-person cases without multi-instance awareness. To address this limitation we introduce 2 new metrics termed Inst-AP and Blink-AP to give attention to both instance awareness quality and eyeblink detection quality.

\textbf{Inst-AP.} It aims to evaluate the instance detection and tracking ability of methods, which is the basis for analyzing the eyeblink behaviors of each instance. We modify the standard evaluation metric video-AP in spatio-temporal action detection~\cite{act_detector} to fit our task. Different from the original implementation, the 3D IoU is calculated between each instance proposal (i.e., a sequence of face bounding boxes) and ground truth across the whole video, rather than the action tube only. As a result, the proposed metric can reveal the algorithm’s ability for detection and tracking instances, while video-AP can just reveal the localization ability of the isolated action tube without instance-level awareness. We 
report the averaged Inst-AP under the IoU from 50\%-95\% with step 5\%. During calculating Inst-AP, each obtained true positive (TP) prediction has matched with one ground-truth instance, and thus, we can calculate the eyeblink detection accuracy of these TP predictions using the proposed Blink-AP. The TP predictions under IoU of 50\% are used to calculate the subsequent Blink-AP.

% The TP predictions under IoU of 50\% will be used to calculate eyeblink detection accuracy (i.e., Blink-AP) as follows.

\textbf{Blink-AP.} This metric is used to reflect the model’s eyeblink detection ability within each instance. We tailor the standard AP metric in temporal action detection~\cite{activitynet} into our task. Only the matched (i.e., true positive) predictions when calculating Inst-AP are taken into consideration in this metric as the blink accuracy of these predictions is rarely affected by the face detection and tracking accuracy. 
% Thus, Blink-AP reveals the pure eyeblink detection ability of the model. 
We report the Blink-AP under the 
temporal IoU of 50\% and 75\%.

% Our proposed metrics consider both the instance-level tracking ability and the action-level eyeblink detection ability of algorithms, which fits for our goals mentioned in~\cref{sec:problem_definition}.

\section{Method}

In this section, we present our InstBlink that takes a video clip as input and directly outputs the face positions of each instance and their eyeblink intervals in the whole video clip. The InstBlink design takes inspiration from the existing query-based methods~\cite{detr,sparsercnn,tuber,tevit,efficientvis} 
% in object detection and video instance segmentation, 
but formulates the architecture to model both face and eyeblink representations at instance level along the video clip. The overall architecture is illustrated in~\cref{fig:pipeline}.

Given a video clip $I\in \mathbb{R} ^{T\times 3 \times H\times W}$ where $T$ denotes the number of frames and $H\times W$ is the spatial size of each frame, InstBlink first applies a backbone network to extract video feature $F_v\in \mathbb{R} ^{T\times C \times H'\times W'}$ where $C$ is the channel number and $H'\times W'$ is the spatial size of the feature. Afterwards, a query-based architecture iterates $M$ times, which consists of 3 components: the Query Interaction Module (QIM), Video Interaction Module (VIM), 
and task-specific heads (i.e., face head and blink head). At the end of each interaction, the queries will be updated and the instance-level face and eyeblink predictions will be output by the task-specific heads. During inference, the output from the last iteration will be used as the final prediction results.

\subsection{Instance Query}
Within InstBlink, the spatio-temporal instance queries $\left\{ q^i \right\} _{i=1}^{N}$ are responsible for characterizing every human instance's joint face and eyeblink features in a video. Each query contains $T$ embeddings (i.e., $q^i \in \mathbb{R}^{T \times C}$), where $C$ is the feature dimension. Each embedding generally focuses on the instance’s face and eyeblink representations in the corresponding frame. Each query is also paired with a proposal tube $p^i \in \mathbb{R}^{T \times 4}$ (i.e., spatio-temporal bounding boxes across the time), which aims at 
indicating the face location of $i$-th instance across the entire video clip. At the first iteration of each complete forward propagation, $q^i$ and $p^i$ are initialized by copying 2 learnable parameters $\bar{q}^i\in  \mathbb{R}^{1\times C}$ and $\bar{p}^i\in  \mathbb{R}^{1\times 4}$ $T$ times along the temporal dimension.

\subsection{Query Interaction Module (QIM)}
QIM targets at (1) enhancing the association between the specifc query and its corresponding human instance and (2) modeling spatio-temporal face and eyeblink representations of the associated instance. Specifically, we first adopt a spatial self-attention layer to allow spatial interaction among queries within each frame:
\begin{equation}
    \left\{\boldsymbol{q}_t^i\right\}_{i=1}^N = \operatorname{MHSA}\left(\left\{\boldsymbol{q}_t^i\right\}_{i=1}^N\right), \quad t \in \left[ 1,T \right],
\end{equation}
where MHSA is the multi-head self-attention~\cite{transformer}. 
Spatial interaction can build strong communication among queries to better model the instance features under complex circumstances such as occlusion due to human interactions.
Then, temporal self-attention is used within each query along the temporal dimension to realize temporal interaction:
\begin{equation}
    \left\{\boldsymbol{q}_t^i\right\}_{t=1}^T = \operatorname{MHSA}\left(\left\{\boldsymbol{q}_t^i\right\}_{t=1}^T\right), \quad i\in \left[ 1,N \right].
\end{equation}
Applying temporal interaction within each query allows the embeddings from different frames to communicate with each other to facilitate instance tracking and model temporal eyeblink representations of the corresponding instance. 

\subsection{Video Interaction Module (VIM)}
VIM aims at collecting the face and eyeblink information of the target instance from the video feature. Particularly, the dynamic filters~\cite{sparsercnn} $m^i$ are first generated from each query embeddings $q^i$. Then, the filters will filter an RoI feature by dynamic convolution to extract highly related features for both face and eyeblink. The RoI feature is obtained by applying RoI align~\cite{maskrcnn} on the video feature using the proposal tube $p^i$.
After obtaining the filtered feature, a linear projection is applied to form the updated query feature $\tilde{q}^i$. The updated query feature will be used to make both face and eyeblink predictions by the task-specific heads.
% After obtaining the filtered feature, we use $\tilde{q}^i=q^i+FC\left( q^i \right)$ to update q, where FC is a linear projection and $\tilde{q}^i$ is the update query feature. $\tilde{q}^i$ will be used to make both face and eyeblink predictions by the task-specific heads.
% Different from the existing works that only extract feature from the pre-extracted local eye region, our blink filter also consider useful global information such as global face appearance, head pose and illumination condition to aid the implicit eye localization and eye state representation. The reason is that the blink filter is derived from the query embeddings where global face context is already stored. During training, the eyeblink information will not only flow back to the RoI feature, but also to the instance query to enhance the eyeblink-related feature representation ability of queries.

\subsection{Task-specific Heads}
The instance-level face and eyeblink predictions can be obtained simultaneously by applying task-specific heads on the query features. The heads are shared among queries.

\noindent \textbf{Face head.}
Given the updated query feature $\tilde{q}^i$, a Multilayer Perceptron (MLP) layer with Sigmoid normalization is used to indicate the existence of human faces as
\begin{equation}
    c^i=Sigmoid\left( MLP_c\left( \tilde{q}^i \right) \right),
\end{equation}
where $c^i\in \mathbb{R} ^{T}$ denotes the face classification scores across frames. Face localization is achieved in a similar way as
% \vspace{-1mm}
\begin{equation}
l^i=MLP_l\left( \tilde{q}^i \right) ,
\end{equation}
% \vspace{-1mm}
where $l^i\in \mathbb{R} ^{T \times 4}$ denotes the face bounding boxes across frames. It will also be used to renew the proposal tube $p^i$.

% (fc + layer norm + relu +fc)and bounding box regression(fc + norm + relu *3  + fc) by separate FFNs.

\begin{figure}[t]
\begin{center}
%\fbox{\rule{0pt}{2in} \rule{0.9\linewidth}{attention.pdf}}
\includegraphics[width=0.48\textwidth]{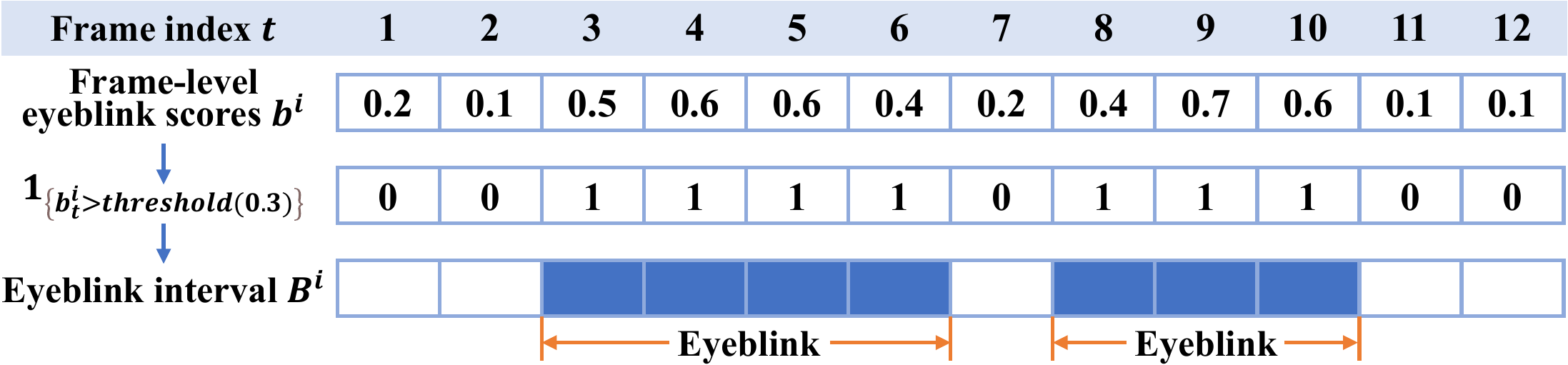}
\vspace{-10.5mm}
\end{center}
        \caption{An example of the blink merging procedure, which takes the frame-level eyeblink scores $b^i$ and outputs the interval-level eyeblink results $B^i$. The threshold is set to 0.3 for demonstration.}
\label{fig:blink_convertor}
\vspace{-5mm}
\end{figure}

\noindent \textbf{Blink head.}
Eyeblink prediction towards $\tilde{q}^i$ is achieved by an MLP layer with Sigmoid normalization as
% \vspace{-1mm}
\begin{equation}
    b^i=Sigmoid\left( MLP_b\left( \tilde{q}^i \right) \right),
\end{equation}
% \vspace{-1mm}
where $b^i\in \mathbb{R} ^{T}$ is the frame-level eyeblink scores. Each element $b^i_t$ in it indicates the possibility that the $t$-th frame is inside an eyeblink event. During inference, we simply merge the adjacent frame-level eyeblink predictions based on a threshold (e.g., 0.3 in our implementation) to form the final interval-level eyeblink predictions $B_k^i=\left[s_k^i, e_k^i\right]$ where $k \in K$ is the $k$-th eyeblink interval within the total K eyeblink predictions. An intuitive example of the merging procedure is shown in~\cref{fig:blink_convertor}.

Different from the existing works that only extract features from the local eye region, our blink head also considers useful global information such as global face appearance, head pose, and illumination condition to aid the implicit eye localization and eyeblink characterization. The reason is that the blink head directly filters global features from the query embeddings where the global face context is already stored. During training, the eyeblink clues will not only flow back to the RoI feature, but also to the instance query to enhance the eyeblink-related feature representation ability of queries. Moreover, because of the shared feature among blink head and face head, a kind of interaction and synergy is built where the information flow from blink head can also benefit the face representations to boost the face detection and tracking ability as verified in~\cref{sec:multi_task}.

As we have $N$ queries, we finally obtain the predictions from them in parallel:
\begin{equation}
    \left\{ y^i \right\} _{i=1}^{N}=\left\{ \left( c^i,l^i,b^i \right) \right\} _{i=1}^{N}.
\end{equation}

\subsection{Training}
To optimize InstBlink, one-to-one assignment between instance-level predictions $\left\{ y^i \right\} _{i=1}^{N}$ and ground truths is first conducted based on their instance-level face tracklet similarity. Then, the loss is calculated concerning face and eyeblink items jointly for optimization. As in~\cref{sec:problem_definition}, the ground truth annotations can be summarized as 
\begin{equation}
    \left\{ \hat{y}^j \right\} _{j=1}^{N_{gt}}=\left\{ \left( \hat{c}^j,\hat{l}^j,\hat{b}^j \right) \right\} _{j=1}^{N_{gt}},
\end{equation}
where $\hat{c}^j\in \mathbb{R} ^{T}$, $\hat{l}^{j}\in \mathbb{R}^{T \times 4}$, and $\hat{b}^j\in \mathbb{R} ^{T}$ indicate the frame-level face existence, face position (i.e., bounding box) and eyeblink existence (i.e., $\hat{b}_t^j = 1$ indicates frame t is inside an eyeblink event and conversely 
for $\hat{b}_t^j = 0$, which can be derived from $\hat{B}^j$ using an inverse process in~\cref{fig:blink_convertor}) respectively. We use Hungarian algorithm~\cite{hungarian} to perform instance-level bipartite matching between predictions and ground truths. The matching cost is given as
\vspace{-5mm}
% \begin{equation}
%     \begin{aligned}
% 	\mathcal{L} _{\mathrm{Hung}}\left( y^i,\hat{y}^j \right) &=\lambda _{cls}\cdot \mathcal{L} _{cls}\left( c^i,\hat{c}^j \right)\\
% 	&+\lambda _{L1}\cdot \mathcal{L} _{L1}\left( l^i,\hat{l}^j \right)\\
% 	&+\lambda _{\mathrm{giou}}\cdot \mathcal{L} _{\mathrm{giou}}\left( l^i,\hat{l}^j \right)\\
% \end{aligned}
% \end{equation}

\begin{equation}
\begin{aligned}
	\mathcal{L} _{\mathrm{Hung}}\left( y^i,\hat{y}^j \right) &=\sum_{t=1}^T\left[\mathcal{L} _{cls}\left( c_t^i,\hat{c_t}^j \right)\right.\\
	&\left.+1_{\left\{ \hat{c}_{t}^{j}\ne 0 \right\}}\left( \mathcal{L} _{\mathrm{box}}\left( l_{t}^{i},\hat{l}_{t}^{j} \right) \right)\right],\\
\end{aligned}
\end{equation}
where $\mathcal{L}_{c l s}$ indicates the focal loss~\cite{focal_loss} for face existence classification.  $\mathcal{L}_{box}$ is the combination of L1 loss and GIoU loss~\cite{giou} for face localization. The loss weights are the same as~\cite{sparsercnn, tevit}. After obtaining the optimal assignment $\hat{\sigma}$ between predictions and ground truths, the network is optimized by the loss as
\vspace{-4mm}

\begin{equation}
\mathcal{L} =\sum_{j=1}^{N_{gt}}{\left( \mathcal{L} _{face}\left( \hat{y}^j,\hat{y}^{\hat{\sigma}\left( j \right)} \right) +\lambda\mathcal{L} _{blink}\left( \hat{y}^j,\hat{y}^{\hat{\sigma}\left( j \right)} \right) \right)},
\end{equation}
% \vspace{-1mm}
where $y^{\hat{\sigma}\left( j \right)}$ is the prediction that has been matched with the ground truth instance $\hat{y}^j$. 
$\mathcal{L} _{face}$ is the face tracklet detection loss that shares the same form as $\mathcal{L} _{Hung}$ (i.e., $\mathcal{L} _{face}\left( \hat{y}^j,\hat{y}^{\hat{\sigma}\left( j \right)} \right) =\mathcal{L} _{Hung}\left( \hat{y}^j,\hat{y}^{\hat{\sigma}\left( j \right)} \right)$), and $\mathcal{L} _{blink}$ denotes instance-level eyeblink detection loss as
$\mathcal{L} _{blink}\left( \hat{y}^j,\hat{y}^{\hat{\sigma}\left( j \right)} \right) =\sum_{t=1}^T{\mathcal{L} _{cls}\left( \hat{b}_t^j,\hat{b}_t^{\hat{\sigma}\left( j \right)} \right)}\,\,$ with $\lambda = 5$.
For unmatched predictions, only $
\sum_{t=1}^T{\mathcal{L} _{cls}\left( c_{t}^{i},\hat{c}_{t}^{i}=0 \right)}
$ is used to supervise their face existence classification output $c^i$ to be close to $0$ (i.e., no face).

\section{Experiment}
\noindent \textbf{Dataset and evaluation metrics.} We first conduct experiments on the proposed MPEblink dataset. We split the dataset into 423 training videos, 128 validation videos, and 135 test videos. The videos from the same movie will only appear in one set. We train the network on the training set, validate it on the validation set and report the performance on the test set. We report Inst-AP to reveal the instance localization ability and Blink-AP to reveal the eyeblink detection ability within instances, as introduced in ~\cref{sec:evaluation_metrics}. Meanwhile, experiments are also conducted on HUST-LEBW for single-person and trimmed cases. Recall, Precision, and F1 score are reported following~\cite{eyeblink_Hu}.

\begin{table}[t]
\setlength\tabcolsep{3pt}
\centering
\scriptsize
% \vspace{-2mm}
\caption{The main results on the MPEblink dataset.}
\vspace{-3mm}
\begin{tabular}{c|c|cc|c}
\toprule
Type                      & Method                                  & Blink-AP$_{50}$     & Blink-AP$_{75}$    & Inst-AP               \\ \midrule
\multirow{2}{*}{Landmark} & Soukupov{\'a} and Cech~\cite{eyeblink_landmark} & 0.50            & 0.05          & \multirow{4}{*}{56.70} \\
                          & Blink detection+~\cite{blinkdetection+}                        & 0.62           & 0.08          &                       \\ \cmidrule{1-4} 
\multirow{3}{*}{Region}   & Hu et al.~\cite{eyeblink_Hu}                               & 2.68           & 0.04          &                       \\
                          & Daza et al.~\cite{eyeblink_alebk}                             & 5.85           & 0.88          &                       \\ \cmidrule{2-5}
                          & InstBlink (Ours)                        & \textbf{27.19} & \textbf{7.16} & \textbf{67.89}  \\ \bottomrule     
\end{tabular}
\label{tab: main result}
\vspace{-5mm}
\end{table}

\begin{table*}[t]
\centering
\scriptsize
\caption{The inference speed comparison on a single NVIDIA 3090 GPU, assuming the compared methods use InsightFace for face detection (time consumption $T$=9.3ms including pre-processing) and landmark detection. \#faces denotes the face amount in the scene.}
% \scriptsize
\vspace{-2mm}
\begin{tabular}{cccccc}
\toprule
Method          & InstBlink (Ours) & Soukupov{\'a} and Cech~\cite{eyeblink_landmark}        & Blink detection+~\cite{blinkdetection+}  & Hu et al.~\cite{eyeblink_Hu}                & Daza et al.~\cite{eyeblink_alebk}              \\ \midrule
Time/image (ms) & \textbf{8.9} + 2.6 for data processing              & $T$(=9.3)+5.4$\times$\#faces & $T$+5.4$\times$\#faces & $T$+5.7$\times$\#faces & $T$+9.1$\times$\#faces \\ \bottomrule
\end{tabular}
\label{tab: running speed}
\vspace{-3.5mm}
\end{table*}

\noindent \textbf{Implementation details.}
We use ResNet-50-FPN~\cite{resnet,fpn} backbone. The network is pre-trained on YouTube-VIS~\cite{youtubevis} under~\cite{tevit} for a general instance representation ability. The query number $N$ and iteration time $M$ are set to 50 and 4 respectively. AdamW~\cite{adamw} optimizer with a batch size of 8 is used to train the model. The initial learning rate is set to 2.5e-5 for the backbone and 2.5e-4 for the other parts. At the training stage, for the sake of memory efficiency, we use a half frame rate ($\sim$12 FPS) for image sampling and the input clip length is set to 11 that is longer than most eyeblink events (i.e., 0.2-0.4s). The frames are also resized to $640\times360$ before sending into the network. The whole training procedure lasts for 10,000 iterations and the learning rate is multiplied by 0.1 at iteration 6000 and 9000. 
During test, the original frame rate is used and the input clip length is set to 36 with a stride of 18. The predictions within adjacent clips are linked via concerning face bounding box IoU.
% During inference, we use the original frame rate of videos to sample images and the input clip length is set to 36 with a stride of 18. We link the predictions between adjacent clips in consideration of the face bounding box IoU.

\subsection{Benchmark Results on MPEblink Dataset}
\noindent \textbf{Baselines.}
Multi-person eyeblink detection in untrimmed videos is a new task that has not been well concerned before. Therefore, we tailor the existing eyeblink detection approaches with a unified instance detection and tracking method. Specifically, we use the state-of-the-art face analysis toolbox InsightFace~\cite{insightface} to achieve face~\cite{scrfd} and landmark~\cite{fake_it_till_you_make_it} detection. To track each instance, we link the face bounding boxes from the adjacent frames based on their similarity on box IoU. After 
obtaining the instance tracklets (i.e., a sequence of face bounding boxes and landmarks), we employ 4 representative eyeblink detection method~\cite{eyeblink_landmark, blinkdetection+, eyeblink_Hu, eyeblink_alebk} within each instance tracklets. Such a sequential pipeline is commonly used in the existing approaches under the single-person assumption.

\noindent \textbf{Main results.} From Table~\ref{tab: main result}, it can be summarized that:

$\bullet$ All of the multi-person eyeblink detection algorithms cannot achieve a satisfactory performance (i.e., Blink-AP$_{50}$ lower than 30\% and Blink-AP$_{75}$ lower than 10\%). This indicates that eyeblink detection under multi-person, untrimmed, and unconstrained scenarios is indeed challenging and has not been well solved yet.

$\bullet$ For Blink-AP, InstBlink significantly outperforms the others by large margins (i.e., 21\% at least of Blink-AP$_{50}$ and 6\% at least of Blink-AP$_{75}$), which verifies the superiority of the proposed framework. We argue that one essential reason is that our framework can model a better long-term temporal eyeblink representation than the frame-based method~\cite{eyeblink_alebk} and the sliding window based methods~\cite{eyeblink_landmark, eyeblink_Hu, blinkdetection+}. Moreover, under the proposed framework, eyeblink features can be facilitated via face’s global context (e.g., head pose and illumination condition) with joint optimization and interaction, while previous works that utilize a sequential manner cannot. From Table~\ref{tab: main result}, it can also be summarized that the landmark-based methods~\cite{eyeblink_landmark,blinkdetection+} perform poorer than the region-based counterparts. We 
think that one essential reason is that landmark detection is unreliable under unconstrained conditions. A similar conclusion has already been made in~\cite{eyeblink_Hu} but the properties of multi-person and long video in MPEblink make it worse for landmark-based eyeblink detection methods. Our method is region-based, but localizes eye region in an implicit way where global face context is included and thus becomes more robust. 

$\bullet$ InstBlink also outperforms others on Inst-AP. We think that is because our framework can better model the long-term spatio-temporal instance representations, while the counterparts achieve tracking under a tracking-by-detection framework, which contains limited spatio-temporal modeling and may suffer from heavy occlusion.

\noindent \textbf{Inference speed analysis.}
 The result is listed in Table~\ref{tab: running speed}, assuming the 4 compared methods use InsightFace for face \& landmark detection and InstBlink inferences within a clip length of 36. It can be seen that InstBlink is also of high inference speed (i.e., 112 FPS for network forwarding) while the real-time capacity of other methods is not superior. As the number of instances increases, their running time also increases while our method is not sensitive to the number of people because of the one-stage inference property.

\begin{table}[]
\setlength\tabcolsep{2pt}
\centering
\scriptsize
\caption{Performance comparison on the HUST-LEBW dataset, where InstBlink\_cross indicates InstBlink trained on MPEblink and tested on HUST-LEBW.}
\vspace{-2.5mm}
\begin{tabular}{c|ccccc}
\toprule
Training   set             & Method                              & Eye   & Recall         & Precision      & F1       \\ \midrule
\multirow{6}{*}{HUST-LEBW~\cite{eyeblink_Hu}} & \multirow{2}{*}{Soukupov{\'a} and Cech~\cite{eyeblink_landmark}} & Left  & 36.07          & 64.71          & 46.32          \\
                           &                                     & Right & 30.16          & 57.58          & 39.58          \\
                           & \multirow{2}{*}{Hu et al.~\cite{eyeblink_Hu}}          & Left  & 54.10           & \textbf{89.19} & 67.35          \\
                           &                                     & Right & 44.44          & 76.71          & 56.28          \\
                           & Blink detection+~\cite{blinkdetection+}                    & Both  & 58.99          & 80.05          & 67.90           \\
                           & InstBlink (Ours)                    & Both  & \textbf{97.64} & 56.62          & 71.68          \\ \midrule
\multirow{3}{*}{mEBAL~\cite{eyeblink_mEBAL_2020}}     & \multirow{2}{*}{Daza et al.~\cite{eyeblink_mEBAL_2020}}        & Left  & 96.03          & 60.80           & 74.46          \\
                           &                                     & Right & 79.50           & 73.48          & 76.37          \\
                           & Daza et al.~\cite{eyeblink_alebk}                         & Both  & 93.39          & 75.33          & 83.39          \\ \midrule
MPEblink                   & InstBlink\_cross (Ours)             & Both  & 91.34          & 76.82          & \textbf{83.45} \\ \bottomrule
\end{tabular}
\vspace{-2mm}
\label{tab: hust-lebw}
\end{table}

\subsection{Benchmark Results on HUST-LEBW Dataset}
Experiments are also conducted on the HUST-LEBW dataset to explore the generality of InstBlink towards single-person and trimmed in-the-wild cases. 
% We copy the coarse video-level eyeblink label to frame-level labels to train Instblink on HUST-LEBW.
% We directly copy the coarse video-level eyeblink label to frame-level labels to supervise InstBlink. 
The results are given in Table~\ref{tab: hust-lebw}. For models trained on the HUST-LEBW dataset, our method still outperforms others by a large margin (i.e., 3.78\% on F1 score), even with limited training data (less than 450 trimmed samples) to train our one-stage model for multiple tasks (face detection, tracking, and eyeblink detection). Meanwhile, the model trained on MPEblink can obtain 83.45\% F1 score on HUST-LEBW. This indeed verifies the strong generalization ability of InstBlink and MPEblink towards eyeblink detection task.
\subsection{Ablation Study}

% \begin{table}[t]
% % \setlength\tabcolsep{3pt}
% \centering
% \scriptsize
% \vspace{-12pt}
% \caption{Ablation studies on QIM and VIM.}
% \vspace{-3mm}
% \begin{tabular}{cccc}
% \Xhline{0.8pt}
% Method                            & Blink-AP$_{50}$     & Blink-AP$_{75}$    & Inst-AP        \\
% \hline
% Full model                        & \textbf{27.19} & \textbf{7.16} & \textbf{67.89} \\
% \hline
% w/o QIM                           & 3.20            & 0.39          & 58.93          \\
% w/o spatial interaction in QIM    & 26.65          & 6.18          & 62.45          \\
% w/o temporal interaction in   QIM & 4.58           & 0.55          & 63.61          \\
% w/o filter operation in VIM       & 22.27          & 4.59          & 65.01          \\
% \Xhline{0.8pt}
% \end{tabular}
% \vspace{-5mm}
% \label{tab: qim_vim}
% \end{table}

\begin{table}[t]
\centering
\scriptsize
\caption{Ablation studies on QIM and VIM.}
\vspace{-3mm}
\begin{tabular}{cccc}
\toprule
Method                            & Blink-AP$_{50}$     & Blink-AP$_{75}$    & Inst-AP        \\
\midrule
w/o QIM                           & 3.20            & 0.39          & 58.93          \\
w/o temporal interaction in QIM & 4.58           & 0.55          & 63.61          \\
w/o spatial interaction in QIM    & 26.65          & 6.18          & 62.45          \\
w/o filter operation in VIM       & 22.27          & 4.59          & 65.01          \\
Full model                        & \textbf{27.19} & \textbf{7.16} & \textbf{67.89} \\
\bottomrule
\end{tabular}
\vspace{-2mm}
\label{tab: qim_vim}
\end{table}

\noindent \textbf{Spatial and temporal modeling in QIM.}
From Table~\ref{tab: qim_vim}, we can see that (1) without applying temporal interaction, the performance of Blink-AP drops from 27.19\% to 4.58\%. This indicates that temporal modeling is critical for eyeblink detection in untrimmed videos (i.e., only using appearance features can not accurately localize eyeblinks). It can also be observed that temporal interaction facilitates the performance of Inst-AP. We argue that this is because it can also build instance-level association among the features across frames to facilitate the instance tracking ability. (2) Spatial interaction can boost Inst-AP by a large margin (i.e., 5.44\%) and slightly boost Blink-AP, as it may build strong communication among queries to better model the instance features under complex circumstances such as occlusion due to human interactions.

\begin{table}[t]
\centering
\caption{The performance comparison with and without blink head on the MPEblink dataset.}
\vspace{-2.5mm}
\scriptsize
\begin{tabular}{cccc}
\toprule
Blink   head & Inst-AP   & Inst-AP$_{50}$ & Inst-AP$_{75}$ \\ \midrule
$\times$            & 65.86          & 81.73          & 71.23          \\
\checkmark            & \textbf{67.89} & \textbf{84.51} & \textbf{73.76} \\ \bottomrule
\end{tabular}
\label{tab: multi-task}
\vspace{-5mm}
\end{table}

\noindent \textbf{Filter operation in VIM.} As in Table~\ref{tab: qim_vim}, compared with directly using RoI features to update query, using filtered RoI features can boost 4.92\% on Blink-AP$_{50}$ and 2.88\% on Inst-AP. We speculate that the filter operation can activate task-specific RoI features and resist background. Thus, finer face and eyeblink clues can be collected to update queries.

\noindent \textbf{Multi-task learning mechanism.} \label{sec:multi_task}
% As the features are shared between the face head and blink head in InstBlink, it can be regarded as a multi-task learning mechanism where the information of different tasks can seamlessly interact. 
Here we study the effect of the blink head from a multi-task learning perspective. As shown in Table~\ref{tab: multi-task}, adding blink head can also boost 2.03\% on Inst-AP, which demonstrates that features for face detection and tracking can also benefit from eyeblink clues.

% \noindent \textbf{Input clip length.} We study the effect of the input clip length during training. As shown in Table, we can observe that: (1) robust (2) length increase - performance increase

\begin{figure}[t]
\begin{center}
%\fbox{\rule{0pt}{2in} \rule{0.9\linewidth}{attention.pdf}}
\includegraphics[width=0.45\textwidth]{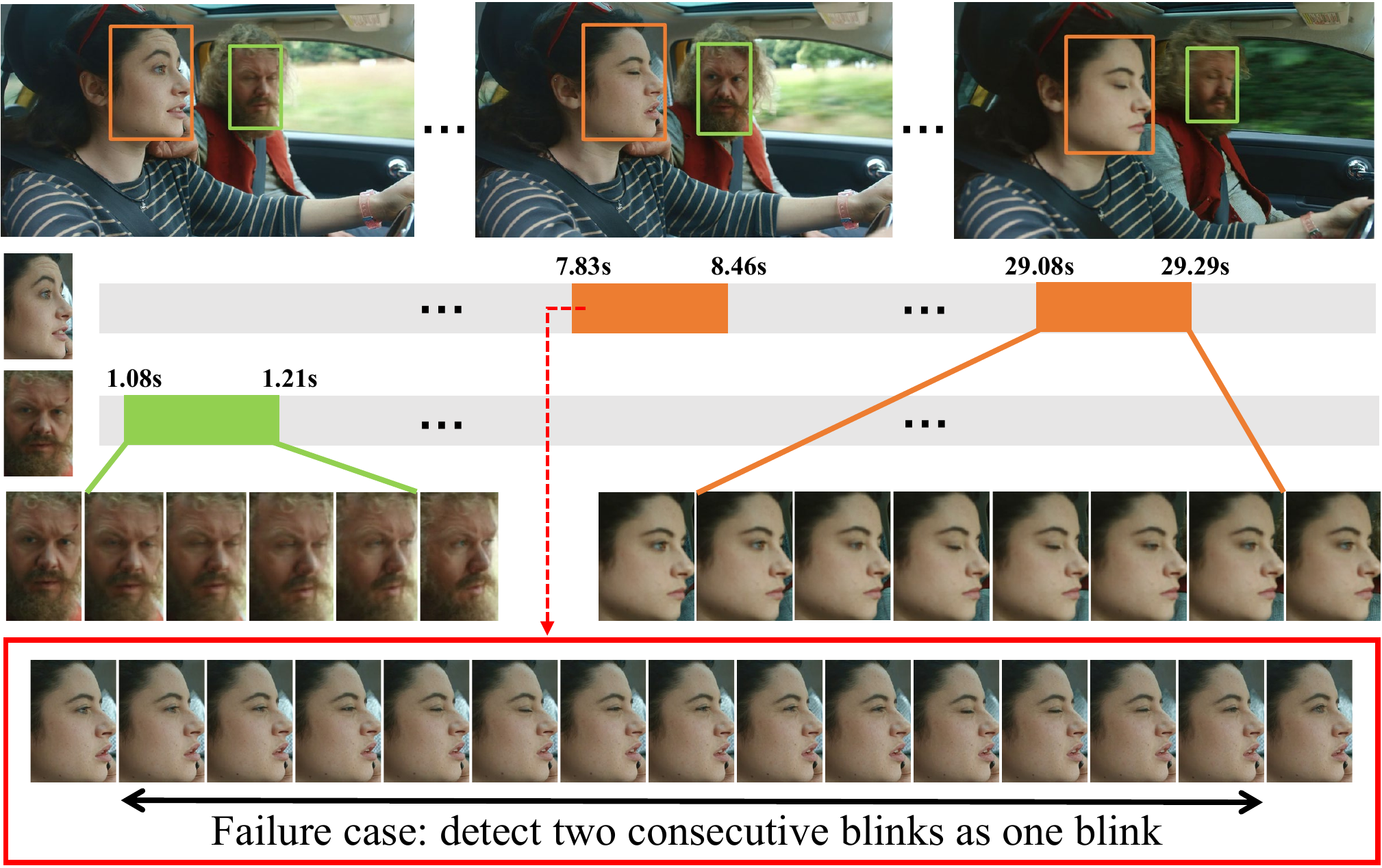}
\vspace{-6mm}
\end{center}
        \caption{Qualitative examples of the predictions of InstBlink.}
\label{fig:visualization}
\vspace{-5mm}
\end{figure}

\subsection{Qualitative Analysis}
We visualize the prediction results of InstBlink in~\cref{fig:visualization}. The examples show that our model can aware the attendees robustly and detect their eyeblinks under different facial appearance and head pose. Nevertheless, it can not distinguish the rapid consecutive eyeblinks, which also reveals the challenge of eyeblink detection in untrimmed videos.

\vspace{-1mm}
\section{Conclusions and Limitations}
\vspace{-1mm}
In this work, we shed the light on a new research task termed multi-person eyeblink detection in the wild for untrimmed video. 
% This new formulation is more realistic to application scenarios and is more challenging. 
We formally define the task and introduce a multi-person eyeblink detection dataset named MPEblink. To perform multi-person eyeblink detection in untrimmed videos, we introduce a one-stage multi-person eyeblink detection framework InstBlink. Experiment results demonstrate the superiority of InstBlink in both effectiveness and efficiency. However, our SOTA performance is still unsatisfactory (i.e., Blink-AP$_{50}$ $\textless$ 30\%). This verifies the challenges of unconstrained multi-person eyeblink detection in long videos. Besides, although the proposed dataset focuses on multi-person, unconstrained, and untrimmed scenarios that are more realistic and have a broader application value, there are lack of crowd scenes (e.g., 
$\textgreater10$ instances) within it. In the future, we will pay more attention to the crowd scenes and enrich the dataset.

\vspace{-1mm}
\section*{Acknowledgment}
\vspace{-1mm}
This work is jointly supported by the National Natural Science Foundation of China (Grant No. 62271221 and U1913602), and Natural Science Foundation of Guangdong Province (Grant No. 2023A1515011260). Joey Tianyi Zhou is funded by the SERC (Science and Engineering Research Council) Central Research Fund (Use-Inspired Basic Research), and the Singapore Government's Research, and Innovation and Enterprise 2020 Plan (Advanced Manufacturing and Engineering Domain) under programmatic Grant A18A1b0045.

%jointly

%%%%%%%%% REFERENCES
{\small
\bibliographystyle{ieee_fullname}
\bibliography{MPEblink}
}

\end{document}